\pdfoutput=1
\documentclass[10pt,twocolumn,letterpaper]{article}

\usepackage{cvpr}
\usepackage{times}
\usepackage{epsfig}
\usepackage{graphicx}
\usepackage{amsmath}
\usepackage{amssymb}

\usepackage{amsthm}
\usepackage{amsfonts}
\usepackage{subfigure}
\usepackage{multirow}
\usepackage{booktabs}
\usepackage{algorithm}
\usepackage{algorithmic}
\usepackage{cite}

\usepackage[breaklinks=true,bookmarks=false]{hyperref}

\cvprfinalcopy 

\def\cvprPaperID{****} 

\begin{document}

\title{Attributes-Guided and Pure-Visual Attention Alignment for Few-Shot Recognition}

\author{Siteng Huang,$^{1, 2}$ Min Zhang,$^{2}$ Yachen Kang,$^{2}$ Donglin Wang$^{2}$\thanks{Corresponding author.}\\
$^1$Zhejiang University\\
$^2$School of Engineering, Westlake University\\
{\tt\small \{huangsiteng, zhangmin, kangyachen, wangdonglin\}@westlake.edu.cn}
}

\maketitle

\begin{abstract}
   The purpose of few-shot recognition is to recognize novel categories with a limited number of labeled examples in each class. To encourage learning from a supplementary view, recent approaches have introduced auxiliary semantic modalities into effective metric-learning frameworks that aim to learn a feature similarity between training samples (support set) and test samples (query set). However, these approaches only augment the representations of samples with available semantics while ignoring the query set, which loses the potential for the improvement and may lead to a shift between the modalities combination and the pure-visual representation. In this paper, we devise an attributes-guided attention module (AGAM) to utilize human-annotated attributes and learn more discriminative features. This plug-and-play module enables visual contents and corresponding attributes to collectively focus on important channels and regions for the support set. And the feature selection is also achieved for query set with only visual information while the attributes are not available. Therefore, representations from both sets are improved in a fine-grained manner. Moreover, an attention alignment mechanism is proposed to distill knowledge from the guidance of attributes to the pure-visual branch for samples without attributes. Extensive experiments and analysis show that our proposed module can significantly improve simple metric-based approaches to achieve state-of-the-art performance on different datasets and settings.
\end{abstract}

\section{Introduction}

The recent success of visual recognition tasks commonly relies on supervised learning from a large number of labeled samples. However, in many practical applications, it is expensive and time-consuming to collect sufficient labeled samples for each category. Inspired by the fact that humans are good at learning to identify objects with very little direct supervision, \textit{few-shot learning} (FSL) has attracted considerable attention. Trained on sufficient labeled samples from known categories (\textit{seen classes}) and given very few labeled samples (\textit{support set}) of a set of new categories (\textit{unseen classes}), few-shot recognition methods aim at classifying unlabeled samples (\textit{query set}) into these new categories. To imitate the process of learning new concepts, seen and unseen classes do not overlap, which makes classical deep learning methods to have generalization issues. Meanwhile, only very few labeled samples are available for the test unseen classes, which may cause severe overfitting when trying common fine-tuning strategies.

\begin{figure}[t]
  \centering
  \includegraphics[width=0.49\textwidth]{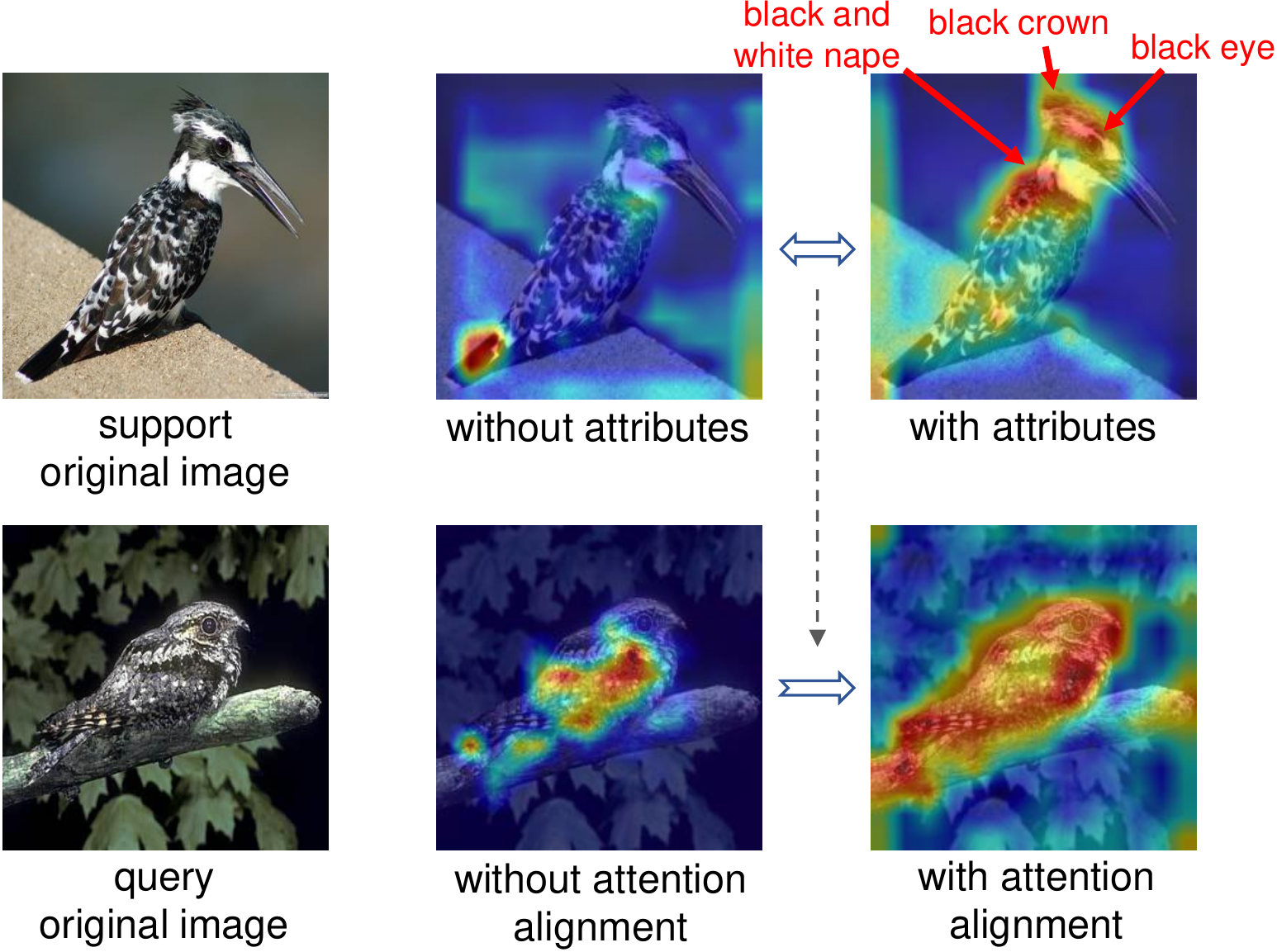}
  \caption{An illustration of the effect of our proposed attention alignment mechanism. The network learns to focus more on discriminative features of support samples with the guidance of auxiliary attributes. And the attention alignment mechanism helps the self-guided branch to learn to select important features without attributes. Best viewed in color.}
  \label{idea_example}
\end{figure}

An effective approach to the few-shot recognition problem is to train a neural network to embed support and query samples into a smaller embedding space, where categories can distinguish with each other based on a distance metric~\cite{Vinyals:MatchNet,Sung:RelationNet}. Existing works have achieved promising results by improving the informativeness and discriminability of the learned representations. Ulteriorly, inspired by the hypothesis that language helps infants to learn to recognize new visual objects~\cite{Jackendoff:linguistic-and-visual}, some recent approaches introduce auxiliary semantic modalities such as label embeddings~\cite{Chen:AM3} and attribute annotations~\cite{Tokmakov:Comp-Feats} to compensate for the lack of supervision. These approaches assume that auxiliary semantic information is only available for support set, but not for query set that is regarded as the prediction object. However, while following this realistic setting, these approaches only focus on the learning of support representations via information mixture or constraint with the help of semantics. The necessity of explicitly designing special mechanisms for query samples has been ignored, resulting in a potential loss of performance. Moreover, as visual and semantic feature spaces naturally have heterogeneous structures, query representations directly obtained from visual contents may shift from same-labeled support representations mixed of both visual and semantic modalities. This is shown as the failure of increasing the intra-class similarity and reducing the inter-class similarity, which damages the accuracy of recognition.

In this paper, we propose a novel attributes-guided attention module (AGAM) to utilize human-annotated attributes as auxiliary semantics and learn more discriminative features. AGAM contains two parallel branches, \textit{i.e.}, the attributes-guided branch and the self-guided branch. Each branch sequentially applies two attention modules, first a channel-wise attention module to blend cross-channel information and learn which channels to focus, then a spatial-wise attention module to learn which areas to focus. The difference between the two branches is that corresponding attributes of support samples can guide the feature selection in the attributes-guided branch, leading to more representative and discriminative representations due to the prominence of relevant elements and noise reduction of irrelevant clutters. And the self-guided branch also helps to refine the pure-visual representations of samples when attributes are not available. Different from existing modality mixture approaches~\cite{Chen:AM3,Schwartz:FSL-MCS} that directly mix multiple modalities with an adaptive proportion, we use the attention mechanism to enhance the informativeness of representations more finely, while ensuring the support representations modified with attributes live in the same space of pure-visual query representations.

Although query representations output by the self-guided branch go through a similar process to support ones, the lack of semantic information may lead to an inaccurate focus on important channels or regions, which increases the distance between same-labeled support and query samples. To handle the issue, we propose an attention alignment mechanism for AGAM, which aligns the attention weights from both branches with a specially-designed attention alignment loss during the learning of support representations. As the features to be emphasized or suppressed by the two branches tend to be similar, the alignment can be regarded as a special case of knowledge distillation~\cite{Hinton:KD}, which means the branch with less information can learn from the branch with more information. Therefore, as shown in Figure~\ref{idea_example}, the self-guided branch can better locate informative features without the guidance of attributes. Note that our AGAM can be viewed as a plug-and-play module, making existing metric-learning approaches more effective. To summarize, our main contributions are in several folds:

1. We utilize powerful channel-wise and spatial-wise attention to learn what information to emphasize or suppress. While considerably improving the representativeness and discriminability of representations in a fine-grained manner, features extracted by both visual contents and corresponding attributes share the same space with pure-visual features.

2. We propose an attention alignment mechanism between the attributes-guided and self-guided branches. The mechanism contributes to learning the query representations by matching the focus of two branches, so that the supervision signal from the attributes-guided branch promotes the self-guided branch to concentrate on more important features even without attributes.

3. We conduct extensive experiments to demonstrate that the performance of various metric-based methods is greatly improved by plugging our light-weight module.

\section{Related Work}

\subsection{Few-Shot Recognition}

Few-shot recognition aims to learn to classify unseen data examples into a set of new categories given only a few labeled samples. Having made significant progress, most meta-learning approaches can be roughly divided into two categories. The first is \textit{optimization-based methods}, which learn a meta-learner to adjust the optimization algorithm so that the model can be good at learning with a few examples, usually by providing the search steps~\cite{Ravi:Meta-Learner-LSTM} or a good initialization to begin the search~\cite{Finn:MAML,Alex:Reptile}. The second is \textit{metric-based methods}~\cite{Vinyals:MatchNet,Snell:ProtoNet,Sung:RelationNet,Oreshkin:TADAM}, which learn a generalizable embedding model to transform all instances into a common metric space, and in this metric space, simple classifiers can be executed directly. For example, Matching Network~\cite{Vinyals:MatchNet} combines both embedding and classification to form an end-to-end differentiable nearest neighbor classifier. Prototypical Network~\cite{Snell:ProtoNet} learns a metric space where embeddings of query samples of one category are close to the centroid (or prototype) of support samples of the same category, and far from centroids of other classes in the episode. Relation Network~\cite{Sung:RelationNet} applies a neural network instead of a fixed nearest-neighbor or linear classifier to evaluate the relationship of each query-support pair.

\subsection{Learning with Semantic Modalities}

With the rapid growth of multimedia data, multimodal analysis has attracted a lot of attention in recent years. In particular, zero-shot learning methods use various semantic modalities to recognize unseen classes without any available labeled samples~\cite{Scott:CUB-splitting,Xian:SUN-splitting}. The common practice in zero-shot learning is to train a projection between visual and semantic feature spaces with labeled samples in seen classes, and apply the learned projection to unseen classes when inferring. Although the setting of zero-shot learning seems similar to that of few-shot learning, simply fine-tuning zero-shot methods with few samples in few-shot problems may lead to overfitting. 

\begin{figure*}[htbp]
  \centering
  \includegraphics[width=0.98\textwidth]{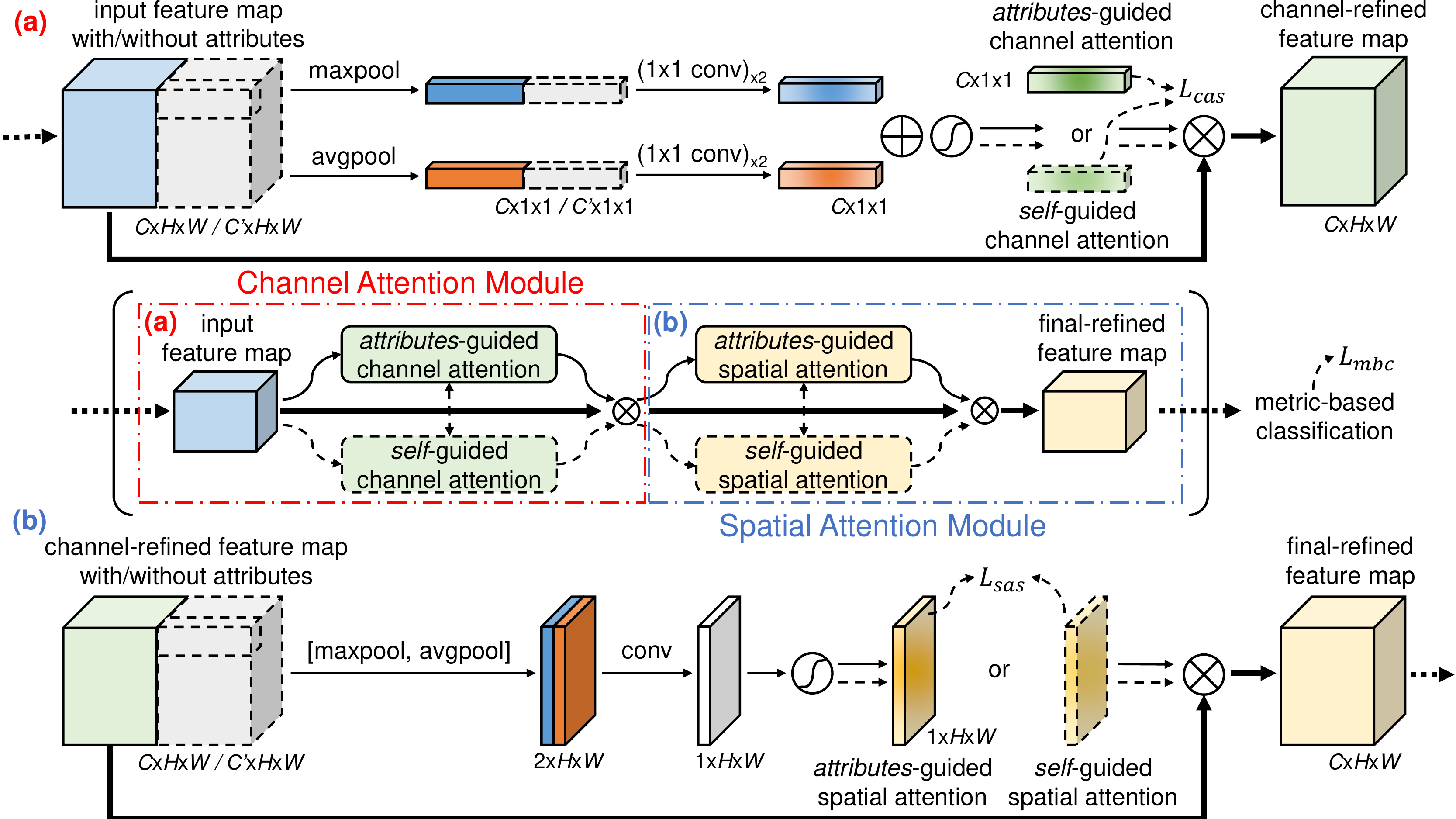}
  \caption{The overall framework of AGAM. Based on whether attributes to the image are available, one of the attributes-guided branch and the self-guided branch is selected. The input features sequentially pass a channel-wise attention module (a) and a spatial-wise attention module (b) to obtain the final-refined features. Best viewed in color.}
  \label{overview}
\end{figure*}

Recently, building upon existing metric-based meta-learning methods, some few-shot learning works propose to utilize auxiliary semantic modalities in a quite different manner from zero-shot learning. \cite{Chen:TriNet} maps samples into a concept space and synthesizes instance features by interpolating among the concepts. \cite{Tokmakov:Comp-Feats} proposes a simple attribute-based regularization approach to learn compositional image representations. \cite{Chen:AM3} models the representation as a convex combination of the two modalities. And \cite{Schwartz:FSL-MCS} proposes a benchmark for few-shot learning with multiple semantics. In our work, with the help of attributes as the only semantic modality, we utilize channel-wise and spatial-wise attention to learn a better metric space in a fine-grained manner. Furthermore, we design an attention alignment mechanism to align the focus of the attributes-guided and self-guided branches, helping to reduce mismatches of same-labeled query and support samples. 

\subsection{Attention Mechanisms}

Motivated by how people focus on different regions of an image or associate words in a sentence, attention mechanisms have been proposed to estimate how strongly one element is correlated with other elements with learned attention vectors. It has been demonstrated that such mechanisms contribute to learning a more informative and robust representation in various areas, such as neural machine translation~\cite{Bahdanau:attention,Vaswani:Transformer}, image processing~\cite{Wang:non-local,Hu:SENet} and generation tasks~\cite{Zhang:SAGAN}. And for few-shot recognition, attention mechanisms are often used to extract more discriminative features. For example, \cite{Ye:FEAT} adapts instance embeddings to task-specific ones with a self-attention architecture. \cite{Chen:CAN} generates cross attention maps for class-query pairs to highlight the target object regions. \cite{Dong:local-FSL} adaptively selects the important local patches among the entire task by introducing episodic attention. And \cite{Yan:dual-attention} applies not only a spatial attention to localize relevant object regions, but also a task attention to select similar training data for label prediction.

\section{Methodology}

\subsection{Preliminaries}

As only a few labeled samples are available in each unseen class, all approaches in our experiments follow the \textit{episodic training} paradigm, which has been demonstrated as an effective approach for few-shot recognition~\cite{Snell:ProtoNet,Sung:RelationNet}. In general, models are trained on $K$-shot $N$-way episodes, and each episode can be seen as an independent task. An episode is created by first randomly sampling $N$ categories from seen classes and then randomly sampling support and query samples from these categories. Our method hypothesizes that both visual contents and attributes as semantic information can be useful for few-shot learning. Therefore, the support set $\mathcal{S} = \left\{\left(s_{i}, a_{i}, y_{i}\right)\right\}_{i=1}^{N \times K}$ contains $K$ labeled examples for each of the $N$ categories. Here, $s_i$ is the $i$-th image, $a_{i}$ denotes the attributes vector to the image, and $y_i \in\{1, \ldots, N\}$ denotes the class label to the image. However, the attributes for query samples are considered to be unavailable, and the query set $\mathcal{Q}=\left\{\left(q_{i}, y_{i}\right)\right\}_{i=1}^{Q}$. Here, $q_i$ is the $i$-th image, and $Q$ denotes the number of query samples. The training phase aims to minimize the loss of the prediction in the query set for each episode, and the performance of the method is measured by the prediction accuracy of new episodes sampling from unseen classes. Note that attributes are not used in some of the experimental comparison approaches. 

\subsection{Algorithm Overview}

In this work, we resort to metric-based methods to obtain proper feature representations for support and query samples, and propose an \textbf{attributes-guided attention module (AGAM)} to modify the features by taking into account the attribute annotations to the images. Figure~\ref{overview} presents an overview of our proposed AGAM. Inspired by~\cite{Woo:CBAM}, we utilize channel-wise attention and spatial-wise attention modules to obtain the final refined features. However, different from the previous work, we design two parallel branches, \textit{i.e.}, \textbf{attributes-guided branch} (denoted by ${ag}$) and \textbf{self-guided branch} (denoted by ${sg}$). For samples with attributes annotations, the attributes-guided branch learns the attention weights by incorporating both attributes and visual contents. And the self-guided branch is designed for the inference of samples without the guidance of attributes. Furthermore, we propose an \textbf{attention alignment mechanism} in AGAM, which aims to pull the focus of the two branches closer, so that the self-guided branch can capture more informative features for query samples without the guidance of attributes. Note that AGAM is a flexible module and can be easily added into any part of convolutional neural networks.

\subsection{Channel-Wise Attention Module} 

Firstly, as each channel of a feature map can be considered as a feature detector~\cite{Zeiler:channel-as-detector}, we produce a 1D channel-wise attention map to focus on ``what'' is meaningful in the given image, as shown in Figure~\ref{overview}(a). Given an intermediate feature map $\mathbf{F} \in \mathbb{R}^{C \times H \times W}$ output by an established convolutional backbone network, based on whether the attributes vector $\mathbf{a} \in \mathbb{R}^D$ corresponding to the original image is available, the input of the channel-wise attention module can be different. As the attributes vector is not available in the self-guided branch, the input $\mathbf{F}_{c\_inp}^{sg}$ is the same as $\mathbf{F}$, where $c\_inp$ denotes the \textit{inp}ut of the \textit{c}hannel attention module. And for samples with attributes, we firstly broadcast $\mathbf{a}$ along height and width dimension of $\mathbf{F}$ to obtain a tensor $\mathbf{A} \in \mathbb{R}^{D \times H \times W}$, then concatenate $\mathbf{F}$ and $\mathbf{A}$ on the channel dimension to get the input of the attributes-guided branch $\mathbf{F}_{c\_inp}^{ag} = \left[\mathbf{F};\mathbf{A}\right] \in \mathbb{R}^{C^{'} \times H \times W}$, where $C^{'} = C + D$ and $\left[;\right]$ denotes the concatenation. 

To compute the channel-wise attention efficiently, max-pooling and average-pooling are first used in parallel to squeeze the spatial dimension of the input feature. As shown in the later ablation study, using both pooling strategies simultaneously can bring complementary and distinctive features. Here we have $\text{MaxPool}(\mathbf{F}_{c\_inp}^{sg}), \text{AvgPool}(\mathbf{F}_{c\_inp}^{sg}) \in \mathbb{R}^{C \times 1 \times 1}$, and $\text{MaxPool}(\mathbf{F}_{c\_inp}^{ag}), \text{AvgPool}(\mathbf{F}_{c\_inp}^{ag}) \in \mathbb{R}^{C^{'} \times 1 \times 1}$. For each branch, features pooled by each pooling layer are then forwarded to an attention generating network, which consists of two convolutions with kernel size 1 and can also be seen as two linear transformations with a ReLU activation in between~\cite{Lin:Network-in-Network}. The purpose of this attention generating network is to generate channel-wise attention after exploiting the inter-channel relationship of features, and note that parameters of this network are not shared between two branches. The element-wise summation is used to merge the results of the same branch. In short, we have

\begin{align}
  \mathbf{M}^{ag}_{c} = &\sigma( \mathbf{W}^{ag}_1 (\mathbf{W}^{ag}_0 (\text{MaxPool}(\mathbf{F}_{c\_inp}^{ag})))
  \notag
  \\&+ \mathbf{W}^{ag}_1 (\mathbf{W}^{ag}_0 (\text{AvgPool}(\mathbf{F}_{c\_inp}^{ag}))) ),    
\end{align}
\begin{align}
  \mathbf{M}^{sg}_{c} = &\sigma( \mathbf{W}^{sg}_1 (\mathbf{W}^{sg}_0 (\text{MaxPool}(\mathbf{F}_{c\_inp}^{sg})))
  \notag
  \\&+ \mathbf{W}^{sg}_1 (\mathbf{W}^{sg}_0 (\text{AvgPool}(\mathbf{F}_{c\_inp}^{sg}))) ),    
\end{align}

\noindent where $\sigma$ denotes the sigmoid activation function, $\mathbf{W}^{ag}_0 \in \mathbb{R}^{(C^{'}/r) \times C^{'}}$, $\mathbf{W}^{ag}_1 \in \mathbb{R}^{C \times (C^{'}/r)}$, $\mathbf{W}^{sg}_0 \in \mathbb{R}^{(C/r) \times C}$, $\mathbf{W}^{sg}_1 \in \mathbb{R}^{C \times (C/r)}$ are parameters of convolutions, and $r$ is a reduction ratio to reduce parameter overhead. Note that the ReLU activation followed by $\mathbf{W}_0$ is omitted for clearer expression. To obtain the channel-refined features, we multiply $\mathbf{M}^{ag}_c, \mathbf{M}^{sg}_c \in \mathbb{R}^{C \times 1 \times 1}$ with the feature map $\mathbf{F}$, expressed as

\begin{align}
  \mathbf{F}_{c\_out}^{ag} = \mathbf{M}^{ag}_c \otimes \mathbf{F}, \quad \mathbf{F}_{c\_out}^{sg} = \mathbf{M}^{sg}_c \otimes \mathbf{F},
\end{align}

\noindent where $\mathbf{F}_{c\_out} \in \mathbb{R}^{C \times H \times W}$ represents the output of the channel-wise attention module in the corresponding branch, and $\otimes$ denotes element-wise multiplication. During multiplication, the channel-wise attention values are broadcasted along the spatial dimension.

\subsection{Spatial-Wise Attention Module} 

As illustrated in Figure~\ref{overview}(b), we also generate a 2D spatial-wise attention map to focus ``where'' is an informative region. The input of the module is $\mathbf{F}_{s\_inp}^{sg} = \mathbf{F}_{c\_out}^{sg} \in \mathbb{R}^{C \times H \times W}$ for the self-guided branch, and $\mathbf{F}_{s\_inp}^{ag} = \left[\mathbf{F}_{c\_out}^{ag};\mathbf{A}\right] \in \mathbb{R}^{C^{'} \times H \times W}$ for the attributes-guided branch. For both two branches, we first apply max-pooling and average-pooling operations along the channel dimension and concatenate the pooled features. Then for each branch, a convolution layer is used to generate the spatial-wise attention map. In short, the attention map is computed as

\begin{align}
  \mathbf{M}^{ag}_{s} = \sigma( f^{ag}(\left[ \text{AvgPool}(\mathbf{F}_{s\_inp}^{ag}); \text{MaxPool}(\mathbf{F}_{s\_inp}^{ag}) \right]) ),
\end{align}
\begin{align}
  \mathbf{M}^{sg}_{s} = \sigma( f^{sg}(\left[ \text{AvgPool}(\mathbf{F}_{s\_inp}^{sg}); \text{MaxPool}(\mathbf{F}_{s\_inp}^{sg}) \right]) ),
\end{align}

\noindent where $\sigma$ denotes the sigmoid activation function. $f$ represents a convolution operation with the filter size of 7$\times$7 and the number of zero-paddings on both sides of 3, whose parameters are also not shared between the two branches. To obtain the final refined features, we multiply $\mathbf{M}^{ag}_s, \mathbf{M}^{sg}_s \in \mathbb{R}^{1 \times H \times W}$ with the channel-refined features in the corresponding branch, which can be expressed briefly as

\begin{align}
  \mathbf{F}_{s\_out}^{ag} = \mathbf{M}^{ag}_s \otimes \mathbf{F}_{c\_out}^{ag}, \quad \mathbf{F}_{s\_out}^{sg} = \mathbf{M}^{sg}_s \otimes \mathbf{F}_{c\_out}^{sg},  
\end{align}

\noindent where $\mathbf{F}_{s\_out} \in \mathbb{R}^{C \times H \times W}$ represents the output of the corresponding branch. During multiplication, we broadcast the spatial-wise attention values along the channel dimension.

\subsection{Attention Alignment Mechanism} 

As AGAM works with other metric-learning approaches, these improved feature embeddings are finally fed into a metric-based learner. For a $K$-shot $N$-way episode containing $Q$ query samples, the metric-based classification loss can be defined as the negative log-probability according to the true class label $y_n \in \{1, 2, \dots, N\}$:

\begin{align}
  L_{mbc} = - \sum^Q_{b=1} \log p (y = y_n | v^q_b),    
\end{align}

\noindent where $v^q_b$ denotes the feature embedding of the $b$-th query sample. Note that $p (y = y_n | v^q_b)$ is the probability of predicting $v^q_b$ as the $n$-th class and can be different in various metric-learning approaches, hence, the specific representation of probability depends on the chosen approach. 

Furthermore, as the lack of attributes annotations may lead the self-guided branch to concentrate on suboptimal features, the metric-based learner is likely to make wrong predictions for query images as the located channels and regions are shifted from those of the same-labeled support samples. Therefore, to encourage the self-guided branch to learn to emphasize or suppress the same features as if attributes have participated in learning, we design an attention alignment mechanism between the attributes-guided branch and the self-guided branch. This is achieved by applying an attention alignment loss to the same type of attention maps obtained from the same support sample but different branches. Among various types of losses that can be used to measure the similarity of attention maps, we choose a soft margin loss according to the experimental results. Specifically, the attention alignment loss from channel-wise and spatial-wise attention maps of the $i$-th sample can be expressed as

\begin{align} 
  l_i^{cas} = \sum_j \log(1 + \exp(-\widetilde{\mathbf{M}}^{ag}_c(j) \otimes \widetilde{\mathbf{M}}^{sg}_c(j))),   \label{softmargin1} 
  \\
  l_i^{sas} = \sum_j \log(1 + \exp(-\widetilde{\mathbf{M}}^{ag}_s(j) \otimes \widetilde{\mathbf{M}}^{sg}_s(j))),   \label{softmargin2}   
\end{align}

\noindent where $\widetilde{\mathbf{M}}$ indicates that the attention map is normalized, and $(j)$ denotes the $j$-th element of the attention map. For each episode, all support samples are taken into account:

\begin{align}
  L_{cas} = \sum^{N*K}_i l_i^{cas}, \quad L_{sas} = \sum^{N*K}_i l_i^{sas}.    
\end{align}

It is noted that our attention alignment mechanism can be regarded as a special case of knowledge distillation~\cite{Hinton:KD}, where attention maps (referred to the ``knowledge'') of the attributes-guided branch (viewed as a teacher model) become the distillation targets for the self-guided branch (viewed as a student model). By mimicking the focusing behaviors of the attributes-guided branch, the self-guided branch with only unimodal input is expected to concentrate on more informative features.

Accordingly, the overall loss of each episode is defined as $L = L_{mbc} + \alpha L_{cas} + \beta L_{sas}$, where $\alpha, \beta$ are the trade-off hyperparameters to balance the effects of different losses. The time complexity of AGAM is $O(C^{'}HW)$, and the space complexity is $O({C^{'}}^2)$. As the complexities vary with the size of the input features, we note that in our experiments, AGAM is inserted after the last convolutional layer of the backbone network to avoid excessive cost.

\section{Experiments}

\subsection{Experimental Setup}

\begin{table*}[htbp]
  \centering
  \resizebox{0.8\linewidth}{!}{
    \begin{tabular}{l|cc||cc}
    \toprule
    \multirow{2}[4]{*}{\textbf{Method}} & \multicolumn{2}{c||}{\textbf{CUB}} & \multicolumn{2}{c}{\textbf{SUN}} \\
\cmidrule{2-5}          & 5-way 1-shot & 5-way 5-shot & 5-way 1-shot & 5-way 5-shot \\
    \midrule
    MatchingNet~\cite{Vinyals:MatchNet}, \textit{paper} & 61.16 $\pm$ 0.89 & 72.86 $\pm$ 0.70 & -     & - \\
    MatchingNet~\cite{Vinyals:MatchNet}, \textit{our implementation} & 62.82 $\pm$ 0.36 & 73.22 $\pm$ 0.23 & 55.72 $\pm$ 0.40 & 76.59 $\pm$ 0.21 \\
    MatchingNet~\cite{Vinyals:MatchNet} \textbf{with AGAM} & \textbf{71.58 $\pm$ 0.30} & \textbf{75.46 $\pm$ 0.28} & \textbf{64.95 $\pm$ 0.35} & \textbf{79.06 $\pm$ 0.19} \\
          & +\textit{8.76} & +\textit{2.24} & +\textit{9.23} & +\textit{2.47} \\
    \midrule
    ProtoNet~\cite{Snell:ProtoNet}, \textit{paper} & 51.31 $\pm$ 0.91 & 70.77 $\pm$ 0.69 & -     & - \\
    ProtoNet~\cite{Snell:ProtoNet}, \textit{our implementation} & 53.01 $\pm$ 0.34 & 71.91 $\pm$ 0.22 & 57.76 $\pm$ 0.29 & 79.27 $\pm$ 0.19 \\
    ProtoNet~\cite{Snell:ProtoNet} \textbf{with AGAM} & \textbf{75.87 $\pm$ 0.29} & \textbf{81.66 $\pm$ 0.25} & \textbf{65.15 $\pm$ 0.31} & \textbf{80.08 $\pm$ 0.21} \\
          & +\textit{22.86} & +\textit{9.75} & +\textit{7.39} & +\textit{0.81} \\
    \midrule
    RelationNet~\cite{Sung:RelationNet}, \textit{paper} & 62.45 $\pm$ 0.98 & 76.11 $\pm$ 0.69 & -     & - \\
    RelationNet~\cite{Sung:RelationNet}, \textit{our implementation} & 58.62 $\pm$ 0.37 & 78.98 $\pm$ 0.24 & 49.58 $\pm$ 0.35 & 76.21 $\pm$ 0.19 \\
    RelationNet~\cite{Sung:RelationNet} \textbf{with AGAM} & \textbf{66.98 $\pm$ 0.31} & \textbf{80.33 $\pm$ 0.40} & \textbf{59.05 $\pm$ 0.32} & \textbf{77.52 $\pm$ 0.18} \\
          & +\textit{8.36} & +\textit{1.35} & +\textit{9.47} & +\textit{1.31} \\
    \bottomrule
    \end{tabular}%
    }
    \caption{Average accuracy (\%) comparison with 95\% confidence intervals before and after incorporating AGAM into existing methods using a Conv-4 backbone. Best results are displayed in \textbf{boldface}, and improvements are displayed in \textit{italics}.}
  \label{tab:adapting}%
\end{table*}%

\textbf{Datasets.} We use two datasets with high-quality attribute annotations to conduct experiments: (1) Caltech-UCSD-Birds 200-2011 (CUB)~\cite{Wah:CUB}, a fine-grained dataset of bird species, containing 11,788 bird images from 200 species and 312 attributes, (2) SUN Attribute Database (SUN)~\cite{Patterson:SUN}, a fine-grained scene recognition dataset, containing 14,340 images from 717 different categories and 102 attributes. To perform meta-validation and meta-test on unseen classes, both datasets are split so that class sets are disjoint. CUB is divided into 100 classes for training, 50 classes for validation, and 50 classes for testing as in~\cite{Chen:CloserLookFSC}. SUN is divided into 580 classes for training, 65 classes for validation, and 72 classes for testing as in~\cite{Xian:SUN-splitting}. Note that we use category-level attributes on the CUB and image-level attributes on the SUN dataset.

\textbf{Experimental Settings.} We experiment with our approach on 5-way 1-shot and 5-way 5-shot settings, and in each episode, 15 query samples per class are used for both training and inference. We report the \textit{average accuracy} (\%) and the corresponding 95\% \textit{confidence interval} over the 10,000 episodes randomly sampled from the test set.

\textbf{Backbone Architectures.} To fairly compare the experimental results, we apply two convolution backbone networks for our proposed AGAM, Conv-4 and ResNet-12, which are commonly used in related works~\cite{Snell:ProtoNet,Finn:MAML,Oreshkin:TADAM}. Conv-4 consists of a stack of four blocks, each of which is a 3$\times$3 convolution with 64 filters followed by batch normalization~\cite{Ioffe:BN}, a ReLU non-linearity, and 2$\times$2 max-pooling, same as in~\cite{Snell:ProtoNet}. ResNet-12 contains four residual blocks and each block has three convolutional layers with 3$\times$3 kernels. The number of filters starts from 64 and is doubled every next block. And at the end of each residual block, a 2$\times$2 max-pooling is applied.

\textbf{Implementation Details.} Our method is trained from scratch and uses the Adam~\cite{Kingma:Adam} optimizer with an initial learning rate $10^{-3}$. Following the settings of~\cite{Chen:CloserLookFSC}, we apply standard data augmentation including random crop, left-right flip, and color jitter in the meta-training stage. And for meta-learning methods, we train 60,000 episodes for 1-shot and 40,000 episodes for 5-shot settings. For AGAM, we set trade-off hyperparameters $\alpha = 1.0$ and $\beta = 0.1$ for all experiments, and each mini-batch contains four episodes. PyTorch~\cite{Paszke:PyTorch} is used to implement our experiments, and we use one NVIDIA Tesla V100 GPU for all experiments. Code is available at \url{https://github.com/bighuang624/AGAM}.

\subsection{Adapting AGAM into Existing Frameworks}

\begin{figure}[htbp] 
  \centering 
  \subfigure[Results on the CUB dataset.]{
  \includegraphics[width=0.48\linewidth]{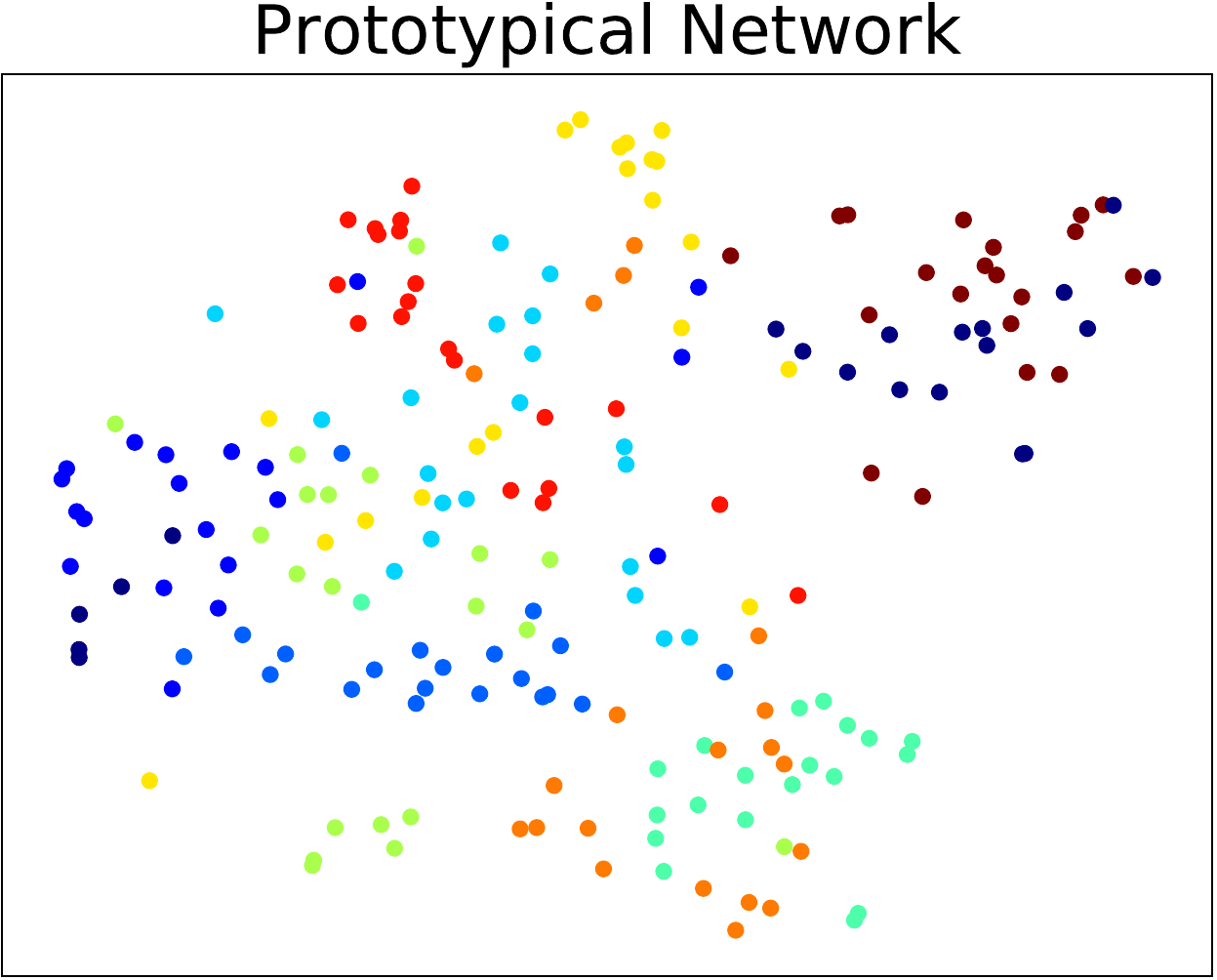} 
  \includegraphics[width=0.48\linewidth]{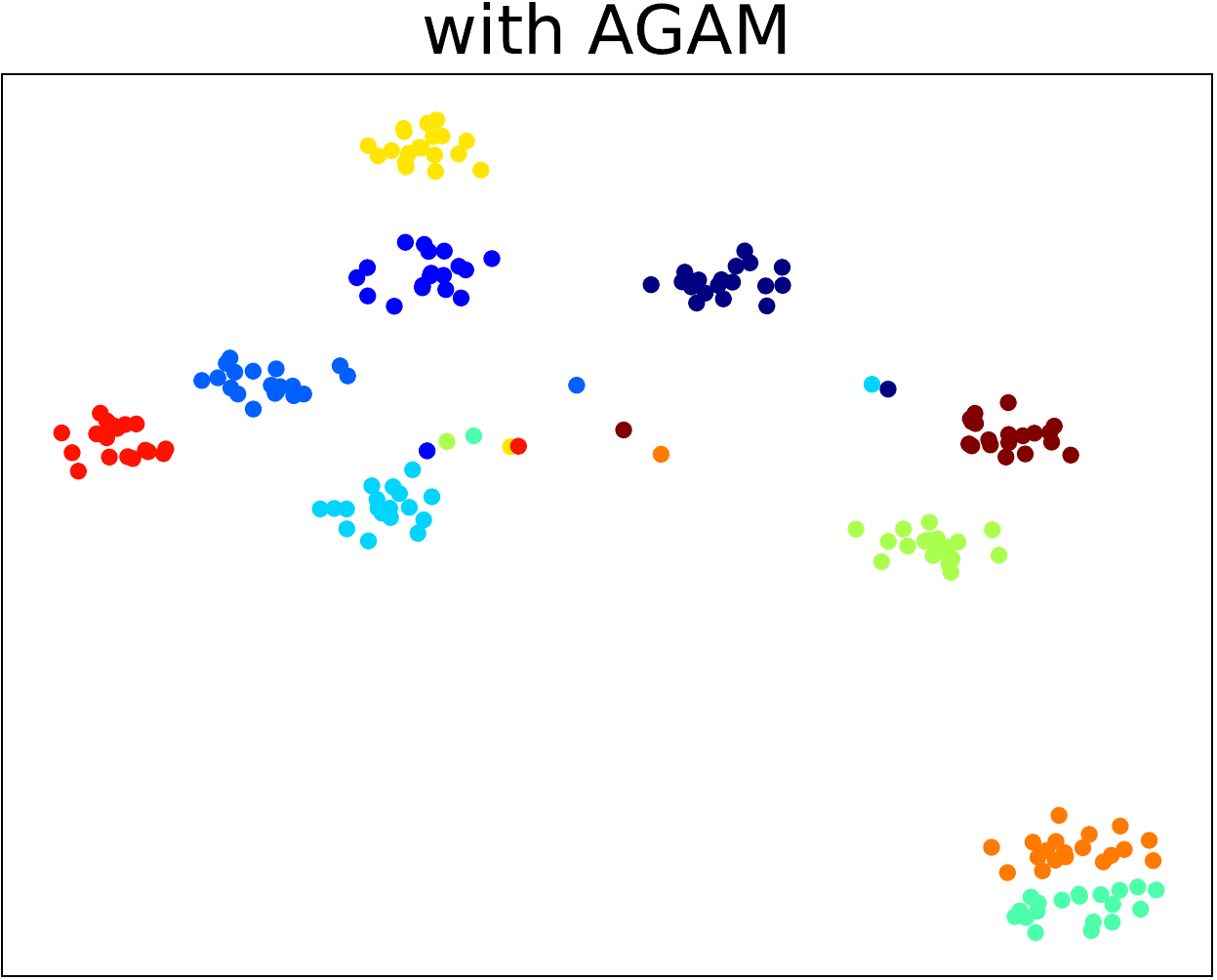} 
  } 
  \subfigure[Results on the SUN dataset.]{
  \includegraphics[width=0.48\linewidth]{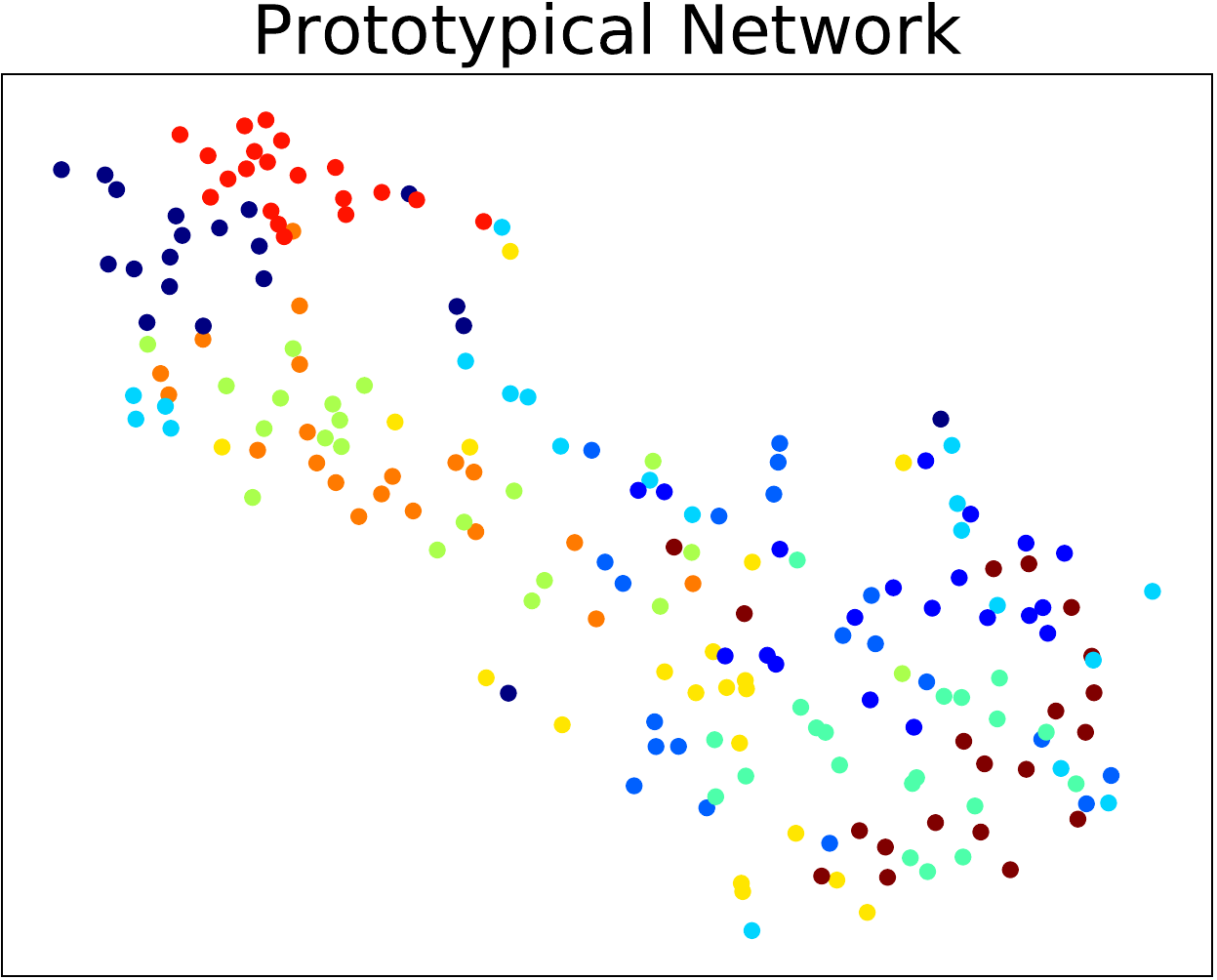} 
  \includegraphics[width=0.48\linewidth]{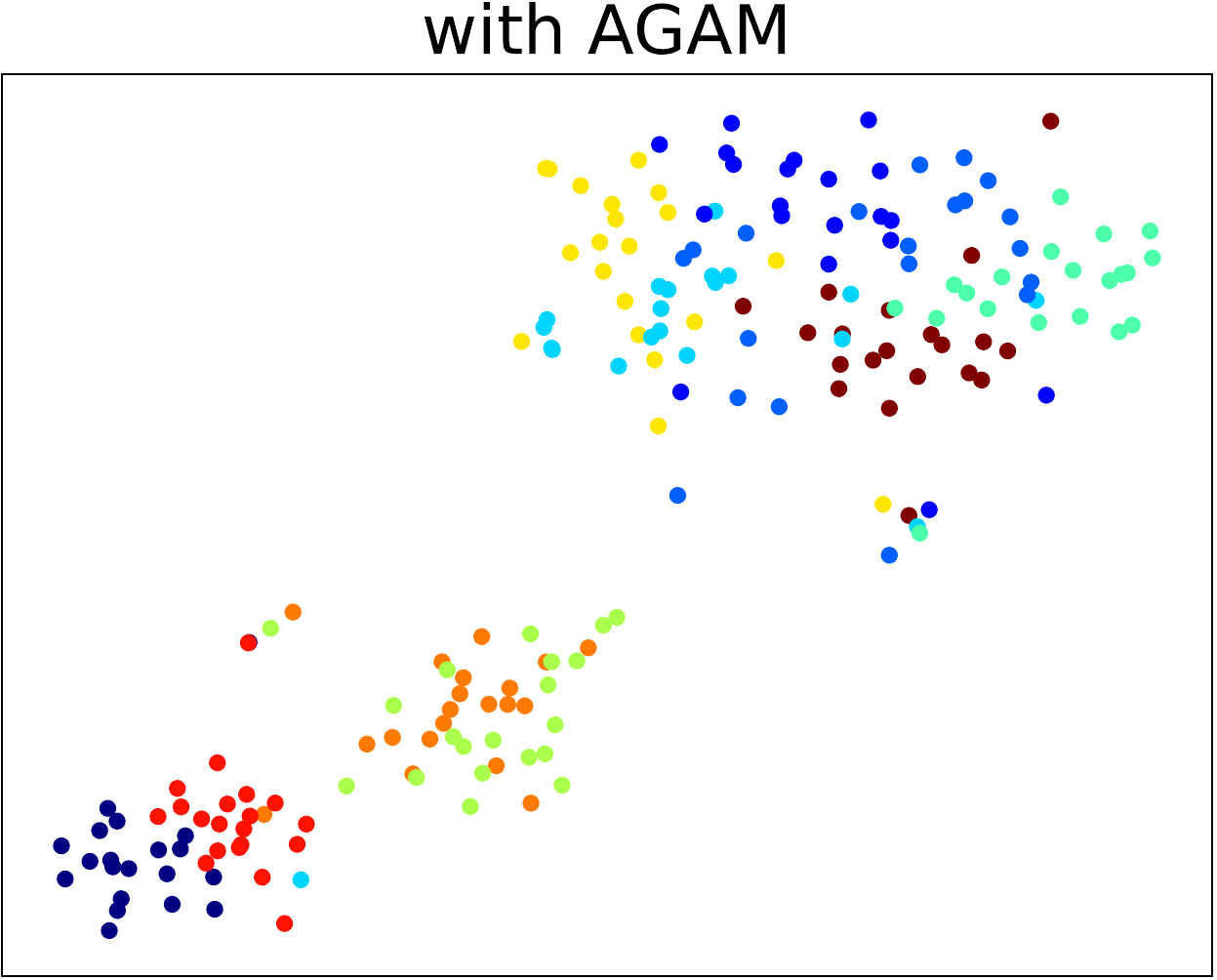} 
  } 
  \caption{The t-SNE visualization of the feature embeddings learned by Prototypical Network with or without our proposed AGAM. Only 10 classes from each dataset are shown for a better view.}
  \label{tSNE}
\end{figure}

\begin{table*}[htbp]
  \centering
  \resizebox{0.62\linewidth}{!}{
    \begin{tabular}{lccc}
    \toprule
    \multirow{2}[2]{*}{\textbf{Method}} & \multirow{2}[2]{*}{\textbf{Backbone}} & \multicolumn{2}{c}{\textbf{Test Accuracy}} \\
          &       & 5-way 1-shot & 5-way 5-shot \\
    \midrule
    MatchingNet~\cite{Vinyals:MatchNet} & Conv-4 & 61.16 $\pm$ 0.89 & 72.86 $\pm$ 0.70 \\
    ProtoNet~\cite{Snell:ProtoNet} & Conv-4 & 51.31 $\pm$ 0.91 & 70.77 $\pm$ 0.69 \\
    RelationNet~\cite{Sung:RelationNet} & Conv-4 & 62.45 $\pm$ 0.98 & 76.11 $\pm$ 0.69 \\
    MACO~\cite{Hilliard:MACO} & Conv-4 & 60.76 & 74.96 \\
    MAML~\cite{Finn:MAML} & Conv-4 & 55.92 $\pm$ 0.95 & 72.09 $\pm$ 0.76 \\
    Baseline~\cite{Chen:CloserLookFSC} & Conv-4 & 47.12 $\pm$ 0.74 & 64.16 $\pm$ 0.71 \\
    Baseline++~\cite{Chen:CloserLookFSC} & Conv-4 & 60.53 $\pm$ 0.83 & 79.34 $\pm$ 0.61 \\
    \midrule
    Comp.~\cite{Tokmakov:Comp-Feats} $^\ast$ & ResNet-10 & 53.6  & 74.6 \\
    AM3~\cite{Chen:AM3} $^\dag$ $^\ast$ & Conv-4 & 73.78 $\pm$ 0.28 & 81.39 $\pm$ 0.26 \\
    AGAM (OURS) $^\ast$ & Conv-4 & \textbf{75.87 $\pm$ 0.29} & \textbf{81.66 $\pm$ 0.25} \\
    \midrule
    \midrule
    MatchingNet~\cite{Vinyals:MatchNet} $^\dag$ & ResNet-12 & 60.96 $\pm$ 0.35 & 77.31 $\pm$ 0.25 \\
    ProtoNet~\cite{Snell:ProtoNet} & ResNet-12 & 68.8  & 76.4 \\
    RelationNet~\cite{Sung:RelationNet} $^\dag$ & ResNet-12 & 60.21 $\pm$ 0.35 & 80.18 $\pm$ 0.25 \\
    TADAM~\cite{Oreshkin:TADAM} & ResNet-12 & 69.2  & 78.6 \\
    FEAT~\cite{Ye:FEAT} & ResNet-12 & 68.87 $\pm$ 0.22 & 82.90 $\pm$ 0.15 \\
    MAML~\cite{Finn:MAML} & ResNet-18 & 69.96 $\pm$ 1.01 & 82.70 $\pm$ 0.65 \\
    Baseline~\cite{Chen:CloserLookFSC} & ResNet-18 & 65.51 $\pm$ 0.87 & 82.85 $\pm$ 0.55 \\
    Baseline++~\cite{Chen:CloserLookFSC} & ResNet-18 & 67.02 $\pm$ 0.90 & 83.58 $\pm$ 0.54 \\
    Delta-encoder~\cite{Bengio:delta-encoder} & ResNet-18 & 69.8  & 82.6 \\
    Dist. ensemble~\cite{Dvornik:Ensemble} & ResNet-18 & 68.7  & 83.5 \\
    SimpleShot~\cite{Wang:SimpleShot} & ResNet-18 & 70.28 & 86.37 \\
    \midrule
    AM3~\cite{Chen:AM3} $^\ast$ & ResNet-12 & 73.6  & 79.9 \\
    Multiple-Semantics~\cite{Schwartz:FSL-MCS} $^\ast$ $^\circ$ $^\bullet$ & DenseNet-121 & 76.1  & 82.9 \\
    Dual TriNet~\cite{Chen:TriNet} $^\ast$ $^\circ$ & ResNet-18 & 69.61 $\pm$ 0.46 & 84.10 $\pm$ 0.35 \\
    AGAM (OURS) $^\ast$ & ResNet-12 & \textbf{79.58 $\pm$ 0.25} & \textbf{87.17 $\pm$ 0.23} \\
    \bottomrule
    \end{tabular}%
  }
  \caption{Average accuracy (\%) comparison to state-of-the-arts with 95\% confidence intervals on the CUB dataset. $^\dag$~denotes that it is our implementation. $^\ast$~denotes that it uses auxiliary attributes. $^\circ$~denotes that it uses auxiliary label embeddings. $^\bullet$~denotes that it uses auxiliary descriptions of the categories. Best results are displayed in \textbf{boldface}.}
  \label{tab:CUB-SOTA}%
\end{table*}%

To verify the effectiveness of our proposed AGAM, we embed it into three metric-based meta-learning approaches: Matching Network~\cite{Vinyals:MatchNet}, Prototypical Network~\cite{Snell:ProtoNet}, and Relation Network~\cite{Sung:RelationNet}. Table~\ref{tab:adapting} shows the gains obtained by incorporating AGAM into each approach on two datasets, and for all three approaches, incorporating AGAM leads to a significant improvement. An observation is that AGAM boosts the performance of Prototypical Network nearly by 23\% on the 1-shot setting and 10\% on the 5-shot setting of CUB, which is especially prominent when compared with other metric-based methods. We believe the reason is that when identifying bird species in such a fine-grained dataset as CUB, a very detailed comparison between the support and the query sample is required. While Matching Network and Relation Network benefit from inputting and analyzing each support-query pair, the original Prototypical Network separately embeds each sample and thus perform worse. However, AGAM supplements the required fine-grained information for Prototypical Network by concatenating on discriminative features, helping to better solve those challenging recognition tasks. 

For the qualitative analysis, we also apply t-SNE~\cite{Maaten:tSNE} to visualize the embedding distributions obtained before and after equipping our AGAM. As shown in Figure~\ref{tSNE}, the model equipped with AGAM has more compact and separable clusters on both datasets, indicating that the learned features are more discriminative due to the guidance of attributes. Moreover, it is observed that in Figure~\ref{tSNE}(a), the effect improvement on the CUB dataset is more significant, which is consistent with the actual accuracy improvement.

\subsection{Comparison with State-of-the-Arts}

\begin{table}[htbp]
  \centering
  \resizebox{\linewidth}{!}{
    \begin{tabular}{lccc}
    \toprule
    \multirow{2}[2]{*}{\textbf{Method}} & \multirow{2}[2]{*}{\textbf{Backbone}} & \multicolumn{2}{c}{\textbf{Test Accuracy}} \\
          &       & 5-way 1-shot & 5-way 5-shot \\
    \midrule
    MatchingNet~\cite{Vinyals:MatchNet} $^\dag$ & Conv-4 & 55.72 $\pm$ 0.40 & 76.59 $\pm$ 0.21 \\
    ProtoNet~\cite{Snell:ProtoNet} $^\dag$ & Conv-4 & 57.76 $\pm$ 0.29 & 79.27 $\pm$ 0.19 \\
    RelationNet~\cite{Sung:RelationNet} $^\dag$ & Conv-4 & 49.58 $\pm$ 0.35 & 76.21 $\pm$ 0.19 \\
    \midrule
    Comp.~\cite{Tokmakov:Comp-Feats} $^\ast$ & ResNet-10 & 45.9  & 67.1 \\
    AM3~\cite{Chen:AM3} $^\dag$ $^\ast$   & Conv-4 & 62.79 $\pm$ 0.32 & 79.69 $\pm$ 0.23 \\
    AGAM (OURS) $^\ast$  & Conv-4 & \textbf{65.15 $\pm$ 0.31} & \textbf{80.08 $\pm$ 0.21} \\
    \bottomrule
    \end{tabular}%
  }
  \caption{Average accuracy (\%) comparison to state-of-the-arts with 95\% confidence intervals on the SUN dataset. $^\dag$~denotes that it is our implementation. $^\ast$~denotes that it uses auxiliary attributes. Best results are displayed in \textbf{boldface}.}
  \label{tab:SUN-SOTA}%
\end{table}%

To prove that simple metric-based methods can surpass the previous state-of-the-art performance after equipping our proposed AGAM, we report the results of our method and others on both CUB and SUN datasets. Note that we choose to display the results of Prototypical Network with our proposed AGAM as our method. For a fair comparison, we split the results achieved by all methods into two groups according to the backbones, and our AGAM uses the same or a smaller backbone network in each group. 

As shown in Table~\ref{tab:CUB-SOTA} and Table~\ref{tab:SUN-SOTA}, our proposed AGAM further improves over Prototypical Network and achieves the best performance among all approaches. It is pointed out that AGAM not only outperforms methods that only use visual contents, but also outperforms methods of utilizing auxiliary semantic information. We attribute this success to two things. The first is that AGAM uses channel-wise and spatial-wise attention to refine the representations from both support and query set in a fine-grained manner, making full use of visual contents and auxiliary attributes. The second is that the attention alignment mechanism matches the focus of two branches, which helps to alleviate the shift between pure-visual query representations and the same-labeled support representations learned with the guidance of attributes.

\subsection{Ablation Study on Using Attributes}

\begin{table}[htbp]
  \centering
  \resizebox{0.94\linewidth}{!}{
    \begin{tabular}{lcc}
    \toprule
    \multirow{2}[2]{*}{\textbf{Method}} & \multicolumn{2}{c}{\textbf{Test Accuracy}} \\
          & 5-way 1-shot & 5-way 5-shot \\
    \midrule
    AGAM  & \textbf{75.87 $\pm$ 0.29} & \textbf{81.66 $\pm$ 0.25} \\
    Not using attributes & 67.35 $\pm$ 0.32 & 77.87 $\pm$ 0.19 \\
    Using all-0 attributes & 63.47 $\pm$ 0.35 & 75.27 $\pm$ 0.20 \\
    ProtoNet~\cite{Snell:ProtoNet} & 53.01 $\pm$ 0.34 & 71.91 $\pm$ 0.22 \\
    \bottomrule
    \end{tabular}%
  }
  \caption{Ablation test results of AGAM on CUB. Average accuracies (\%) with 95\% confidence intervals of each model are reported. Best results are displayed in \textbf{boldface}.}
  \label{tab:ablation_attributes}%
\end{table}%

To prove that attribute annotations do contribute to the improvement of the performance, we construct ablation experiments about attributes in two ways. Specifically, we do one of the following operations at a time: (1) Not use attribute annotations. (2) Set all attributes as 0. We evaluate all these models on the CUB dataset based on Prototypical Network with a Conv-4 backbone, and the results are reported in Table~\ref{tab:ablation_attributes}. It is observed that (1) Only using the proposed attention mechanism can refine the features and lead to better performance. (2) Setting attributes as all-0 vector is better than Prototypical Network but worse than not using attributes, indicating that all-0 attributes may cause damage to attention modules. (3) Using attribute annotations with complete AGAM outperforms the above two methods by a large margin.

\subsection{Ablation Study on Framework Design}

To empirically show the effectiveness of our framework design, we do one of the following operations at a time: (1) Exchange the order of two attention modules. (2) Remove one of the two pooling layers in both branches. (3) Remove one of the two attention modules in both branches. (4) Remove one or both $L_{cas}$ and $L_{sas}$. The ablation results are shown in Table~\ref{tab:ablation_study}.

\begin{table}[htbp]
  \centering
  \resizebox{0.84\linewidth}{!}{
    \begin{tabular}{lcc}
    \toprule
    \multirow{2}[2]{*}{\textbf{Method}} & \multicolumn{2}{c}{\textbf{Test Accuracy}} \\
          & 5-way 1-shot & 5-way 5-shot \\
    \midrule
    AGAM  & \textbf{75.87 $\pm$ 0.29} & \textbf{81.66 $\pm$ 0.25} \\
    AGAM\_SACA & 74.22 $\pm$ 0.27 & 79.72 $\pm$ 0.26 \\
    w/o avgpool & 66.27 $\pm$ 0.29 & 76.58 $\pm$ 0.25 \\
    w/o maxpool & 67.60 $\pm$ 0.29 & 77.09 $\pm$ 0.22 \\
    w/o CA & 54.91 $\pm$ 0.36 & 80.52 $\pm$ 0.24 \\
    w/o SA & 69.66 $\pm$ 0.31 & 76.24 $\pm$ 0.27 \\
    w/o $L_{cas}$ & 74.88 $\pm$ 0.26 & 77.78 $\pm$ 0.26 \\
    w/o $L_{sas}$ & 74.29 $\pm$ 0.27 & 77.87 $\pm$ 0.23 \\
    w/o $L_{cas} \& L_{sas}$ & 75.37 $\pm$ 0.31 & 78.92 $\pm$ 0.27 \\
    \bottomrule
    \end{tabular}%
  }
  \caption{Ablation test results of AGAM on CUB. Average accuracies (\%) with 95\% confidence intervals of each model are reported. Best results are displayed in \textbf{boldface}.}
  \label{tab:ablation_study}%
\end{table}%

\textbf{Influence of the Order of Attention Modules.} We first exchange the order of channel-wise attention and spatial-wise attention in AGAM, with spatial-wise attention in the front and channel-wise attention in the back (\textbf{AGAM\_SACA}). This leads to about 2\% performance drops on both 1-shot and 5-shot settings. The results show that the design of the module sequence is effective.

\textbf{Influence of Pooling Layers.} Removing either of the average-pooling (\textbf{w/o avgpool}) and the max-pooling (\textbf{w/o maxpool}) leads to 8\% to 9\% performance drops on the 1-shot setting and 4\% to 5\% performance drops on the 5-shot setting. This demonstrates that using both pooling strategies simultaneously captures more useful information.

\textbf{Influence of Attention Modules.} Removing the channel-wise attention (\textbf{w/o CA}) leads to about 1\% performance drops on the 5-shot setting, and more than 20\% drops on the 1-shot setting. This fully demonstrates the importance of channel-wise attention in our proposed AGAM when labeled data is particularly scarce. Removing the spatial-wise attention (\textbf{w/o SA}) drops the performance by 5\% to 6\% on both 1-shot and 5-shot settings, which attests that the spatial-wise attention brings stable and significant improvement.

\textbf{Influence of Attention Alignment Loss.} To prove the effectiveness of our proposed attention alignment mechanism, we also remove one or both $L_{cas}$ and $L_{sas}$. An observation from the results is that removing both $L_{cas}$ and $L_{sas}$ (\textbf{w/o $L_{cas} \& L_{sas}$}) obtains about 1\% performance higher than removing one of them alone (\textbf{w/o $L_{cas}$}, \textbf{w/o $L_{sas}$}) on both 1-shot and 5-shot settings. We believe the reason is that when only aligning weights from the channel-wise or spatial-wise attention modules, the important features of query samples can not be consistently selected using only visual information in the other attention module, and therefore it is better to use neither $L_{cas}$ nor $L_{sas}$. However, compared to the complete AGAM model, all three models still perform less than 1\% worse on the 1-shot setting and 2\% to 3\% worse on the 5-shot setting, demonstrating that the joint use of two attention alignment loss items can lead to improvement.

\subsection{Ablation Study on Attention Alignment Loss}

\begin{table}[htbp]
  \centering
  \resizebox{0.76\linewidth}{!}{
    \begin{tabular}{lcc}
    \toprule
    \multirow{2}[2]{*}{\textbf{Loss Type}} & \multicolumn{2}{c}{\textbf{CUB}} \\
 & 5-way 1-shot & 5-way 5-shot \\
    \midrule
    L1    & 66.95 $\pm$ 0.30 & 78.40 $\pm$ 0.25 \\
    MSE   & 69.83 $\pm$ 0.30 & 77.35 $\pm$ 0.22 \\
    smoothL1 & 72.42 $\pm$ 0.30 & 75.72 $\pm$ 0.31 \\
    soft margin & \textbf{75.87 $\pm$ 0.29} & \textbf{81.66 $\pm$ 0.25} \\
    \midrule
    \midrule
    \multirow{2}[2]{*}{\textbf{Loss Type}} & \multicolumn{2}{c}{\textbf{SUN}} \\
 & 5-way 1-shot & 5-way 5-shot \\
    \midrule
    L1    & 60.56 $\pm$ 0.33 & 76.14 $\pm$ 0.26 \\
    MSE   & 59.54 $\pm$ 0.35 & 78.35 $\pm$ 0.26 \\
    smoothL1 & 62.07 $\pm$ 0.31 & 78.42 $\pm$ 0.23 \\
    soft margin & \textbf{65.15 $\pm$ 0.31} & \textbf{80.08 $\pm$ 0.21} \\
    \bottomrule
    \end{tabular}%
  }
  \caption{Ablation test results of different attention alignment losses based on AGAM with a Conv-4 backbone. Average accuracies (\%) with 95\% confidence intervals of each model are reported. Best results are displayed in \textbf{boldface}.}
  \label{tab:loss-type}%
\end{table}%

To prove that the soft margin loss is an ideal choice for the attention alignment mechanism, we compare the results on CUB and SUN with four different candidate losses: (1) \textbf{L1}: measuring the mean absolute error (MAE) between each element in the pair of attention maps. (2) \textbf{MSE}: measuring the mean squared error (squared L2 norm) between each element in the pair of attention maps. (3) \textbf{smoothL1}: a squared term if the absolute element-wise error falls below 1 and an L1 term otherwise, less sensitive to outliers than the MSE loss. (4) \textbf{soft margin}: A deformation of the logistic loss between each element in the pair of attention maps, as expressed in Eq.~\ref{softmargin1} and~\ref{softmargin2}.

The results are reported in Table~\ref{tab:loss-type}. On both CUB and SUN datasets, choosing the soft margin loss in the attention alignment mechanism brings in better performance.

\subsection{Analysis of Attention Alignment Mechanism}

\begin{figure}[h]
  \centering
  \includegraphics[width=0.48\textwidth]{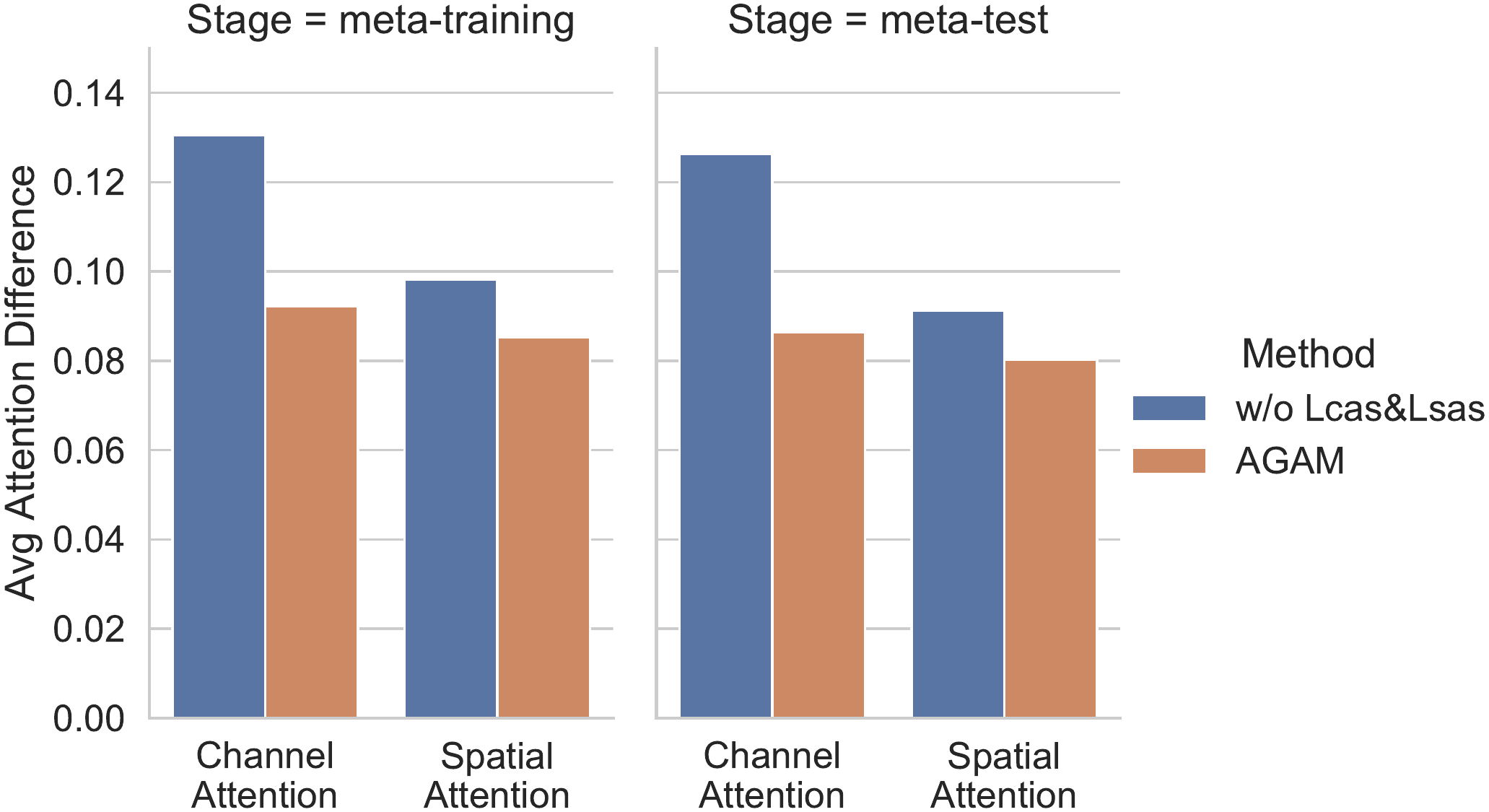}
  \caption{Comparison of average attention difference between the two branches. The results are obtained with AGAM with or without attention alignment losses on meta-training and meta-test episodes. Best viewed in color.}
  \label{AttnDiff}
\end{figure}

To quantify the effect of our attention alignment mechanism on the attention weights generated by the two branches, Figure~\ref{AttnDiff} shows the average attention difference between the two branches when retaining or removing both $L_{cas}$ and $L_{sas}$. We can find that the attention alignment mechanism does contribute to the reduction of the difference between the two branches, demonstrating that the self-guided branch becomes more inclined to focus on the same features as the attributes-guided branch.

\subsection{Grad-CAM Visualization Analysis}

\begin{figure}[h]
  \centering
  \includegraphics[width=0.48\textwidth]{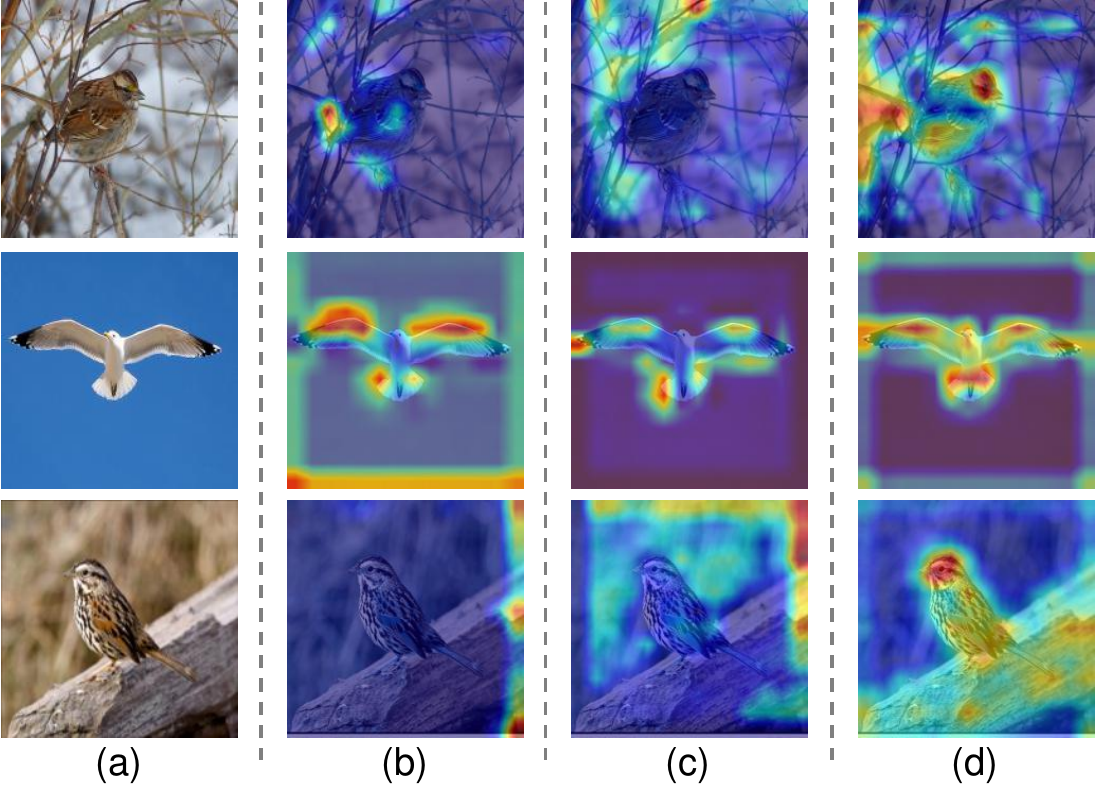}
  \caption{Gradient-weighted class activation mapping (Grad-CAM) visualization of query samples. Each row is the result of the same query sample, and each column is: (a) Original images. (b) Results of Prototypical Network. (c) Results of AGAM but removing the attention alignment mechanism. (d) Results of the complete AGAM. Best viewed in color.}
  \label{GradCAM}
\end{figure}

To qualitatively evaluate whether AGAM indeed exploits the important features with the help of the attention alignment mechanism, Figure~\ref{GradCAM} visualizes the gradient-weighted class activation maps~\cite{Selvaraju:GradCAM} from Prototypical Network, with AGAM but removing the attention alignment mechanism, and with the complete AGAM. It is observed that incorporating AGAM helps to attend to more representative local features than the original Prototypical Network, which contributes to better recognition performance. Also, the masks of AGAM-integrated model cover the representative regions better for query samples when using the attention alignment mechanism, indicating that the self-guided branch benefits from the attention alignment mechanism.

\section{Conclusion}

In this paper, we propose an attributes-guided attention module (AGAM) to fully utilize manually-encoded attributes in few-shot recognition. Owing to the channel-wise attention and spatial-wise attention, the enhanced representations are more unique and discriminative for both support and query samples. Furthermore, through the well-designed attention alignment mechanism, attention alignment is achieved between the attributes-guided branch and the self-guided branch, which narrows the gap between the representations learned with and without attributes. We have demonstrated that our proposed AGAM boosts the performance of metric-based meta-learning approaches by a large margin, which are superior to that of state-of-the-arts.

\section*{Acknowledgments}

The authors gratefully acknowledge funding support from the Westlake University and Bright Dream Joint Institute for Intelligent Robotics.

{\small
\bibliographystyle{ieee_fullname}
\bibliography{ms}
}

\end{document}